\documentclass[sigconf, table, dvipsnames]{acmart}
\usepackage{bbm}
\usepackage{bm}
\usepackage{booktabs}
\usepackage{multirow}
\usepackage{caption}
\usepackage{subcaption}
\usepackage{amsmath}
\usepackage[export]{adjustbox}
\usepackage{xfrac}
\newcommand{\mat}{\bm}
\usepackage{tcolorbox}
\tcbuselibrary{breakable}
\usepackage{enumitem}
\AtBeginDocument{%
  \providecommand\BibTeX{{%
    \normalfont B\kern-0.5em{\scshape i\kern-0.25em b}\kern-0.8em\TeX}}}

\setcopyright{acmcopyright}
\copyrightyear{2023} 
\acmYear{2023} 
\setcopyright{rightsretained} 
\acmConference[DocEng '23]{ACM Symposium on Document Engineering 2023}{August 22--25, 2023}{Limerick, Ireland} 
\acmBooktitle{ACM Symposium on Document Engineering 2023 (DocEng '23), August 22--25, 2023, Limerick, Ireland}
\acmDOI{10.1145/3573128.3609344} 
\acmISBN{979-8-4007-0027-9/23/08}
\begin{document}

\title{Improving Zero-Shot Text Matching for Financial Auditing with Large Language Models}
% \title{Large Language Models for Financial Auditing}


\author{Lars Hillebrand}
\orcid{0000-0002-5496-4177}
\email{lars.patrick.hillebrand@iais.fraunhofer.de}
\authornote{Both authors contributed equally to this research.}
\author{Armin Berger}
\authornotemark[1]
\author{Tobias Deußer}
\authornote{The authors are also affiliated with the University of Bonn in Germany.}

\affiliation{%
  \institution{Fraunhofer IAIS}
  \city{Sankt Augustin}
  \country{Germany}
}

% \author{Thiago Bell}
% \author{Robin Stenzel}
% \author{Milad Morad}
% \author{Christian Temath}
% \affiliation{%
%   \institution{Fraunhofer IAIS}
%   \city{Bonn}
%   \country{Germany}
% }

% \author{Tim Dilmaghani}
% \author{Bernd Kliem}
% \author{Rüdiger Loitz}
% \author{Rafet Sifa\textsuperscript{\textdagger}}
% \affiliation{%
%   \institution{Fraunhofer IAIS\textsuperscript{\textdagger} \\ PricewaterhouseCoopers GmbH}
%   \city{Bonn\textsuperscript{\textdagger} and Düsseldorf}
%   \country{Germany}
% }

\author{Tim Dilmaghani}
\author{Mohamed Khaled}
\author{Bernd Kliem}
\author{Rüdiger Loitz}
\affiliation{%
  \institution{PricewaterhouseCoopers GmbH}
  \city{Düsseldorf}
  \country{Germany}
}

\author{Maren Pielka}
\author{David Leonhard}
\author{Christian Bauckhage}
\authornotemark[2]
\author{Rafet Sifa}
\authornotemark[2]
\affiliation{%
  \institution{Fraunhofer IAIS}
  \city{Sankt Augustin}
  \country{Germany}
}

\renewcommand{\shortauthors}{Hillebrand and Berger, et al.}  % Anonymous Authors

\begin{abstract}

Auditing financial documents is a very tedious and time-consuming process. As of today, it can already be simplified by employing AI-based solutions to recommend relevant text passages from a report for each legal requirement of rigorous accounting standards. However, these methods need to be fine-tuned regularly, and they require abundant annotated data, which is often lacking in industrial environments. Hence, we present ZeroShotALI, a novel recommender system that leverages a state-of-the-art large language model (LLM) in conjunction with a domain-specifically optimized transformer-based text-matching solution. We find that a two-step approach of first retrieving a number of best matching document sections per legal requirement with a custom BERT-based model and second filtering these selections using an LLM yields significant performance improvements over existing approaches.

% Natural language processing methods have several applications in automated auditing, including document or passage classification, information retrieval, and question answering. However, training such models requires a large amount of annotated data which is scarce in industrial settings. At the same time, techniques like zero-shot and unsupervised learning allow for application of models pre-trained using general domain data to unseen domains.
% In this work, we study the efficiency of unsupervised text matching using Sentence-Bert, a transformer-based model, by applying it to the semantic similarity of financial passages. Experimental results show that this model is robust to documents from in- and out-of-domain data.
%    We present a novel pipeline

% In automated auditing, natural language processing techniques have diverse uses such as categorizing documents or passages, retrieving information, and answering questions. However, training these models necessitates abundant annotated data, which is often lacking in industrial environments. Nevertheless, methods like zero-shot learning and unsupervised learning enable the utilization of pre-trained models that have been trained on general data for domains they haven't encountered before.

% This study focuses on examining the effectiveness of unsupervised text matching in financial passages using Sentence-Bert, a transformer-based model. Through experimentation, the results demonstrate that this model proves to be efficient in achieving semantic similarity in the context of financial texts.
\end{abstract}



% \keywords{text mining, natural language processing, recommender system, sustainability reporting, climate change}
%, Large Language Models, Text Matching}

% A "teaser" image appears between the author and affiliation
% information and the body of the document and typically spans the
% page.
% \centering
%   \includegraphics[width=0.8\textwidth]{figures/sustain_screenshot.png}
%   \caption{A screenshot of the sustain.AI recommender tool. After selecting a specific regulatory requirement from one of the categories, the system predicts the most relevant segments of a provided sustainability report. On the right side, the recommended segments are highlighted in the rendered report, fostering an efficient sustainability analysis.}
%   % \Description{Enjoying the baseball game from the third-base
%   % seats. Ichiro Suzuki preparing to bat.}
%   \label{fig:sustain_screenshot}


\maketitle

\section{Introduction}

% On an annual basis large companies are forced to disclose their economic and financial situation
% Corporate disclosure documents like annual financial statements enable 

The annual disclosure of corporate financial statements plays a vital role in informing the public about a company's financial situation and future prospects. The published documents contain detailed information about its financial stability, productivity, and profitability and thus influence external investor's decisions. 
% and profitability and thus influence investment decisions made by external investors. 
Due to their economic significance, these documents are highly regulated. On an annual basis, trained auditors thoroughly introspect and proofread the mandatory disclosures and ensure their compliance according to the legal requirements of the applicable accounting standard, e.g. IFRS (International Financial Reporting Standards). This manual examination process requires high-grade expert knowledge and experience and proves to be inherently time-consuming and prone to human error.
% While requiring a great amount of expert knowledge and experience, this manual examination process still proves to be inherently time-consuming and prone to human error. 
The regulatory requirements of IFRS and other accounting standards are generally presented as a large collection of individual checklist items. For each item, 
% of these items,
the assigned auditor has to identify the relevant text segments in the financial report before answering the
% actual
completeness and correctness questions raised by the requirement. %The first necessity of matching relevant sections in the disclosed financial statement to individual regulatory requirements of the accounting standard
The first retrieval task is particularly tedious, considering the report size and the number of items in the standard.
% (around a thousand legal requirements for IFRS)
%For a regulatory framework like IFRS, encompassing around a thousand legal requirements and financial reports spanning several hundred pages, the described retrieval task consumes a significant portion of the auditor's time.

% Financial statements are documents that contain financial information of organisations such as assets, liabilities, revenues. These documents are examined annually to check conformity with the relevant financial reporting framework, such as International Financial Reports Standards (IFRS) and Handelsgesetzbuch (HGB). The examination process requires a lot of expert knowledge and manual analysis of lengthy financial texts. It includes tasks such as verifying the completeness, accuracy, valuation, consistency, classification, and understandability of the reported information. 

To alleviate this tedious process, we have previously introduced a tool called Automated List Inspection (ALI) \cite{Sifa19}, a recommender system linking paragraphs in a financial document to their corresponding requirements.
% This paper extends our previous work, which developed a suite of machine learning-based applications trained on domain knowledge, called Automated List Inspection (ALI). 
In light of the recent advances of LLMs 
% that have shown remarkable performance on various downstream tasks
, we aim to examine whether domain-specific solutions are still required or can be augmented by LLMs in such a recommender context. Therefore, we compare different architectures that leverage GPT-4 \cite{OpenAI23}, a state-of-the-art LLM,
% (Generative Pre-Trained Transformer 4, introduced in \cite{OpenAI23}), 
for the task of auditing financial documents. Considering the heterogeneous performance gains of GPT-4 reported in OpenAI’s technical report, we investigate the potential and limitations of LLMs in the domain of financial auditing. 
In this work, we focus on the task of matching relevant text segments from financial statements to concrete legal requirements from an accounting standard. 

Therefore, our contributions are introducing a novel method, called ZeroShotALI, which enables zero-shot text matching between new financial reports and unseen legal requirements based on a pretrained SentenceBERT \cite{reimers-2019-sentence-bert} model from \cite{biesner2022zeroshot} and GPT-4 \cite{OpenAI23}, and we evaluate multiple strong baselines, e.g. combining a vector store-based architecture utilizing OpenAI's Ada embeddings with GPT-4.

% In the following, we first review related work, before describing our modeling approach in Section~\ref{section:methodology}. In Section~\ref{section:experiments}, we outline our dataset, present our experiments, and discuss the results. Section~\ref{section:conclusion} then concludes and adds an outlook into conceivable future work.

\section{Related Work}
% Erstes ALI Paper: \cite{Sifa19},
% AliBERT: \cite{Ramamurthy21},
% General recommender: \cite{biesner2022zeroshot},
% KPI Extraction / Numerical Consistency Check: \cite{hillebrand2022kpi}, \cite{hillebrand2022towards}, \cite{deusser2022kpiedgar}, \cite{ali2023automatic}
% Report Generation: \cite{generating_financial_reports_from_tabular_data},
% Contradiction Detection: \cite{deusserCont2023}

To capture the field of related work, we briefly look at two areas of research: The use of natural language processing (NLP) in the financial domain and the use of OpenAI's GPT models in particular.

% The field of automated auditing and
% , with it, the larger field of financial NLP has been of interest to researchers for several years.
The large field of financial NLP and the subfield of automated auditing have been of interest to researchers for several years.
In 2019 we introduced the Automated List Inspection (ALI) tool \cite{Sifa19}, a supervised recommender system that ranks textual components of financial documents according to the requirements of established regulatory frameworks, such as IFRS.
% The field of automated auditing and, with it, the larger field of financial natural language processing has been of interest to researchers for several years. \cite{Sifa19} introduced the Automated List Inspection (ALI) tool, a supervised recommender system that ranks textual components of financial documents according to the requirements of established regulatory frameworks, such as  IFRS.
% To accomplish the ranking task, the authors employed traditional NLP techniques like Tf-Idf, latent semantic indexing, neural networks, and logistic regression. 
% The combination of the first and last methods yielded the best performance. 
Subsequently, in \cite{Ramamurthy21}, we enhanced ALI by utilizing a pre-trained BERT language model developed by \cite{Devlin19} to encode text segments. With \cite{biesner2022zeroshot}, we introduced a more general framework for this task. Regarding a more detailed method for extracting information pertaining to automatic consistency checks of financial disclosures, in \cite{hillebrand2022towards}, we introduced KPI-Check. This BERT-based system utilizes a customized model for named entity and relation extraction, as presented in our previous work, \cite{hillebrand2022kpi}, to automatically identify and validate semantically equivalent key performance indicators in financial reports. Similarly, in \cite{deusser2022kpiedgar}, we studied the KPI extraction task on an English dataset which was released together with the results.
% In a subsequent study, \cite{Ramamurthy21} enhanced ALI by utilizing a pre-trained BERT language model developed by \cite{Devlin19} to encode text segments. Furthermore, \cite{biesner2022zeroshot} introduced a more general framework for this task. Regarding a more detailed method for extracting information pertaining to automatic consistency checks of financial disclosures, \cite{hillebrand2022towards} introduced KPI-Check. This BERT-based system utilizes a customized model for named entity and relation extraction, as presented in \cite{hillebrand2022kpi}, to automatically identify and validate semantically equivalent key performance indicators in financial reports. In a similar vein, \cite{deusser2022kpiedgar} studied the KPI extraction task on an English dataset which they released together with their results.
% There is work on fraudulent statement detection by \cite{temponeras19}, who implemented a Deep Dense Multilayer Perceptron and compared with other models such as Decision Trees, k-Nearest Neighbours, Support Vector Machines, and Logistic Regression, as well as \cite{zhu2021novel}, who put forth a novel capsule network to detect fraudulent activities in accounting reports. 
% Within \cite{ali2023automatic} we investigated how a plethora of pre-trained tabular models can check the consistency of tables and text in financial reports. 
Within \cite{ali2023automatic} we investigated how a plethora of pre-trained tabular models can check the table and text consistency in financial reports. 
% In their study, \cite{cao2018towards} employed a combined entity and relation extraction method to verify formulas in Chinese financial documents. 
% Yet another important aspect of financial auditing is checking for any form of contradictions in annual reports.
Another important aspect of financial auditing is the detection of contradictions in annual reports. In \cite{deusserCont2023} we studied how such a task can be automated with a transformer-based model. 
As for the implementation of entire LLMs specialized in financial language, \cite{wu23bloomberggpt} introduced \textsc{BloombergGPT}, employing Bloomberg's extensive data sources and evaluating the model's performance on financial tasks and general LLM benchmarks. 
% \cite{ali2023automatic} investigated how a plethora of pre-trained tabular models can check the consistency of tables and text in financial reports. 
% % In their study, \cite{cao2018towards} employed a combined entity and relation extraction method to verify formulas in Chinese financial documents. 
% Yet another important aspect of financial auditing is checking for any form of contradictions in annual reports. \cite{deusserCont2023} studied how such a task can be automated with a transformer-based model. 
% As for the implementation of entire LLMs specialized in financial language, \cite{wu23bloomberggpt} introduced \textsc{BloombergGPT}, employing Bloomberg's extensive data sources and evaluating the model's performance on financial tasks and general LLM benchmarks. 
% Another LLM, FinBERT, was introduced by \cite{araci2019finbert} and further employed and researched by \cite{yang2020finbert}, \cite{liu2021finbert}, \cite{arslan2021comparison}, and \cite{huang2022finbert}.

As the GPT line of OpenAI is still relatively novel, research exploring its capabilities and limitations is still conducted. However some results have already been found in various domains. When it comes to finance, \cite{cao2023bridging} demonstrated GPT-4's effectiveness in sentiment analysis, ESG analysis, corporate culture analysis, and Federal Reserve opinion analysis qualitatively using practical examples. 
Quantitative analysis was carried out by \cite{neilson2023artificial} who used ChatGPT to create financial recommendations for the Australian financial sector, finding, that ChatGPT failed to operate effectively with complex financial advise, requiring additional professional guidance. 

\section{Methodology}
\label{section:methodology}

% In this section, we formally define the problem of matching text segments within documents to relevant legal requirements before turning to the in-depth analysis of our proposed architecture which is visualized in Figure \ref{fig:architecture}.

\begin{figure}[t]
  \centering
  \includegraphics[width=0.75\linewidth]{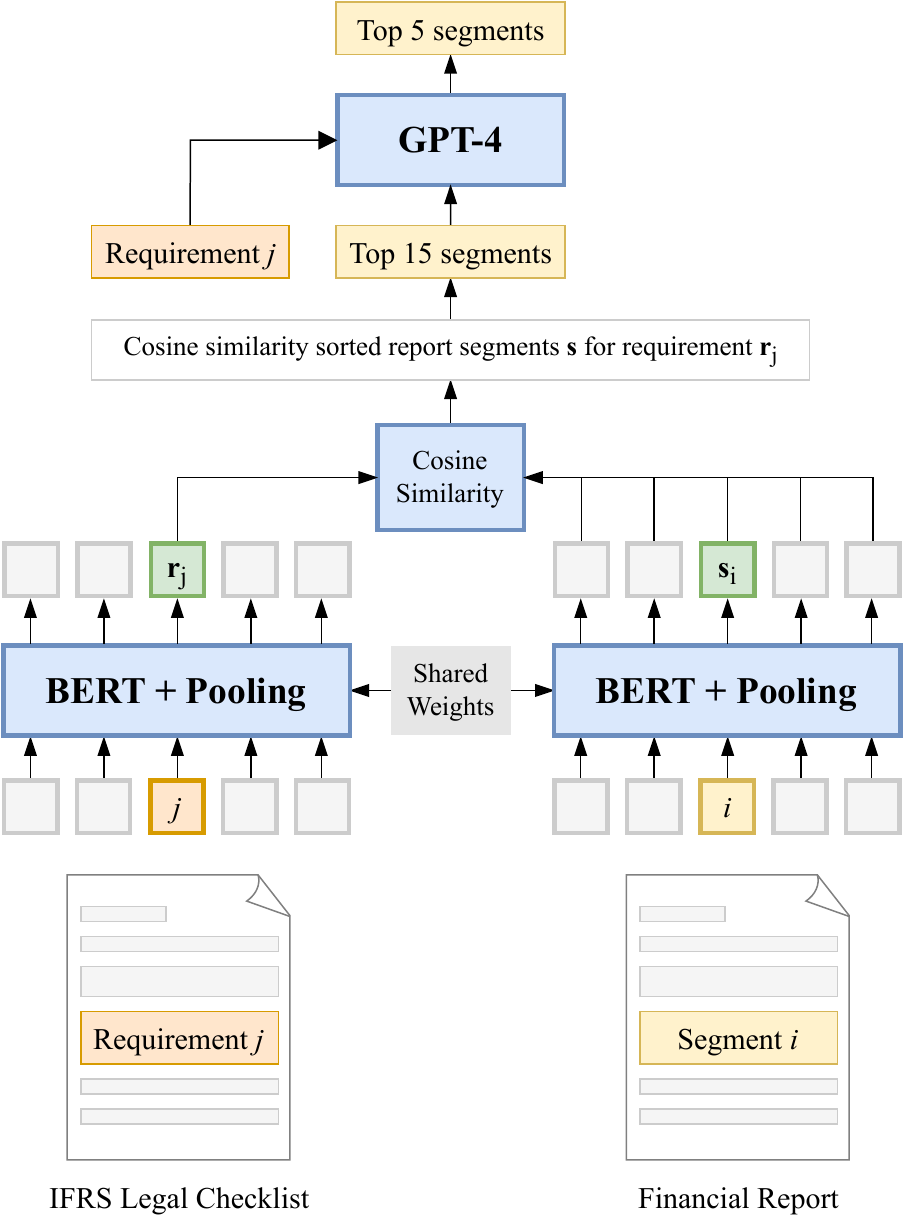}
  \caption{Schematic visualization of the ZeroShotALI recommender system. A domain-adapted SentenceBERT model matches relevant financial report segments $\mat{s}$ to concrete legal requirements $\mat{r}$ based on cosine similarity before GPT-4 further enhances the final matching outcome.}
  \label{fig:architecture}
\end{figure}

Formally, the task of assigning relevant text segments from a financial report to concrete requirements from an accounting standard can be defined as a text matching problem: For each legal requirement $r_{j} \in \mathcal{R}$, a list of relevant text segments from the report $s_{i} \in \mathcal{S}$ (where $\mathcal{S}$ is the set of all paragraphs and $\mathcal{R}$ is the set of all requirements) has to be predicted. The recommendation system has to assign relevance scores $n_{i,j} \in (0, 1)$ to every segment-requirement pair $(s_{i}, r_{j})$. To obtain a classification from those scores, the top $K$ text segments for each requirement are being selected as relevant.
% Formally, the task of assigning relevant text segments from a financial report to concrete requirements from an accounting standard can be defined as a text matching problem, in which for each legal requirement $r_{j} \in \mathcal{R}$, a list of relevant text segments from the report $s_{i} \in \mathcal{S}$ (where $\mathcal{S}$ is the set of all paragraphs in a report, and $\mathcal{R}$ is the set of all requirements) has to be predicted. So, the recommendation system has to assign relevance scores $n_{i,j} \in (0, 1)$ to every segment-requirement pair $(s_{i}, r_{j})$. To obtain a classification from those scores, the top $K$ text segments for each requirement are being selected as relevant.

Previous work has mostly approached this task as a multi-label classification problem, where a sigmoidal output layer predicts individual relevance scores for each pre-defined legal requirement. This problem formulation is inflexible with respect to unseen requirements, i.e. a full model re-training is required. Also, the semantic information contained in the actual requirement texts is neglected, since the requirements have been translated to numeric class ids.
% Previous work has mostly approached this matching task by means of a multi-label classification problem, where a sigmoidal output layer predicts individual relevance scores for each pre-defined legal requirement. This problem formulation is inflexible with respect to unseen legal requirements, i.e. a full model re-training is required. In addition, the semantic information contained in the actual requirement texts is neglected, since the requirements have been translated to numeric class ids.

To overcome these architectural shortcomings, we introduce ZeroShotALI. Following \cite{biesner2022zeroshot}, we employ a text similarity based matching model that individually encodes text segments from the financial report as well as legal requirements from the accounting standard before predicting matches based on semantic text similarity. Specifically, we leverage a domain-adapted SentenceBERT model as our initial text retrieval solution. It is a modification of the BERT language model \cite{Devlin19} that 
% uses the transformer architecture \cite{vaswani2017attention}. SentenceBERT 
encodes both the requirement text and the financial report text segment, using two BERT models with shared weights, and applies mean-pooling to obtain paragraph level embeddings $\mat{s}_i$ for text segment $i$ and $\mat{r}_j$ for requirement $j$, respectively. Subsequently, the cosine similarity between these embeddings is computed to measure how well the report segment and legal requirement semantically match, resulting in a normalized similarity score between 0 and 1. For full architectural details and training procedures, we refer to \cite{biesner2022zeroshot}. To further improve the final matching performance, we augment the described SentenceBERT model with the state-of-the-art generative language model, GPT-4. In a first pre-filtering step, SentenceBERT retrieves the top 15 most relevant financial report segments for each requirement query $j$. Then, we prompt GPT-4 with these 15 segments and the requirement text as input to further narrow down the recommendations to the five best matching segments. The complete architecture is illustrated in Figure \ref{fig:architecture}. We evaluate and compare the two-stage approach of ZeroShotALI with multiple baselines in the next section.

\section{Experiments}
\label{section:experiments}

In the following, we introduce our custom data set, describe competing baseline methods and different LLM prompt designs, as well as evaluation metrics. Finally, we also discuss the results we obtained from our approaches.
% In the following section, we introduce our custom data set, describe competing baseline methods and different LLM prompt designs, as well as evaluation metrics. Finally, we also discuss the results we obtained from our approaches.

% In the following sections, we introduce our two custom data sets of German sustainability reports, define our evaluation metrics, discuss the overall training setup, describe the competing baseline methods, and finally, evaluate results.
\subsection{Data}
\label{data}

We obtained 10 IFRS-compliant reports from PwC as the data source for this paper, containing a total of 7097 text segments. The reports are labeled by auditors, who map the text segments in each report to one of the corresponding 1214 IFRS requirements. This process was done in multiple iterations by different domain experts, in order to assure a high annotation quality. Since, except for the SentenceBERT model, none of the four architectures require additional training, we only employ a test set to compare the different architectures. The fine-tuned SentenceBERT model has not been trained or evaluated on the employed test set to prevent data leakage. %We convert each report into a JSON format containing text segments with unique ids.% 
For more information on the domain data utilized to fine-tune the SentenceBERT model, we direct the reader to \cite{biesner2022zeroshot}.

\subsection{Baselines}
\label{baselines}

We compare ZeroShotALI with (1) a plain SentenceBERT model as described in Section \ref{section:methodology} without the GPT-4 filtering, (2) a vector database retrieval model employing OpenAI's Ada V1 or V2 embeddings, and (3) the same vector database approach combined with GPT-3.5 Turbo or GPT-4 as filtering models.

The vector database we use is Chroma DB. This open-source vector database combines a k-means clustering algorithm with the ClickHouse database management system to retrieve semantically similar text passages given an input query. For more details on the concept and an overview of similar applications, we refer to \cite{Aumuller20}. We embedded each of the 10 IFRS reports using either the lower dimensional Ada V1 or Ada V2 embeddings and stored them in the database separately so that each report could be accessed by its name, and we could query the system based on a specific report. For each query, we retrieved the top five most semantically similar text segments using their cosine similarity score.

The third set of systems consists of a vector database, Chroma DB, using V2 Embeddings and frozen LLMs, GPT-3.5 Turbo or GPT-4. These systems follow the idea of retrieval-augmented generation, which aims to provide LLMs with relevant information for a specific task instead of relying solely on the language model’s parametric knowledge. As before, we retrieved the top 15 most relevant text segments for each query using Chroma DB. Then, we used these 15 segments as prompt input for the LLM, which was then tasked to return the five most relevant text segments for each requirement.

\subsection{Effects of prompt design}

For all architectures leveraging GPT-4 we evaluate the impact of prompt design on model performance. Prompt design refers to the format and phrasing of the task presented to the GPT-style LLMs. Building upon the findings of \cite{Liu21} regarding the effects of prompt phrasing on LLM performance, we conduct our evaluations by primarily investigating two factors: (1) the phrasing of the task and (2) the structure of the output allowed for the model's responses.

Our evaluation is as follows: We formulate a specific task for GPT-4, which involves the retrieval of the five most relevant text segments from a provided set of 15 segments, based on a given requirement. We then randomly sample 20 requirements from one of the 10 IFRS-compliant reports. %The performance of the model is measured using a custom performance metric called ``one-shot recall'', which we elaborate on in the Evaluation Metric section. 
Regarding the phrasing of the task, we find no significant impact on the quality of the model's responses, implying that variations in the way the task was presented did not substantially influence the model's performance.
% To evaluate the effects of prompt design, we formulate a specific task for GPT-4, which involves the retrieval of the five most relevant text segments from a provided set of 15 segments, based on a given requirement. We randomly sample 20 requirements from one of the 10 IFRS-compliant reports obtained from our partner, PwC Germany. %The performance of the model is measured using a custom performance metric called ``one-shot recall'', which we elaborate on in the Evaluation Metric section. 
% Regarding the phrasing of the task, we find no significant impact on the quality of the model's responses. It implies that variations in the way the task was presented did not substantially influence the model's performance.
However, when examining the structure of the output allowed for the model's responses, we observe a notable effect. We categorize the types of outputs into two broad formats: ``open-ended'' and ``closed.'' In the ``open-ended'' format, the model is permitted to provide explanations for its answers, while in the ``closed'' format, the model is restricted to returning the IDs of the five most relevant text segments. Interestingly, we find that the ``closed'' format yields improved performance compared to the ``open-ended'' format. Table \ref{tab:prompt_comparison} shows the four best performing prompts and their respective performance.\footnote{Due to the stochastic nature of GPT-4 it is impossible to exactly reproduce results. We set the temperature parameter to $0$, which reduces but does not remove stochasticity in the generation process.} For all GPT-based experiments we leverage the best performing prompt design A.

%Due to the stochastic nature of GPT-4\footnote{We set the temperature parameter to $0$, which reduces but does not remove stochasticity in the generation process.}, it is challenging to precisely reproduce results, and it is anticipated that there will be performance variations over time.
\begin{table}
\caption{Quantitative comparison of 4 different prompt setups 
%to determine the best GPT Prompt 
for the text segment to requirement matching task. Reported metrics are defined in Section \ref{sec:metrics}. The {\color{NavyBlue}{(bracketed blue)}} text shows the respective differences between prompts B and A as well as D and C.}
\centering
\begin{tabular}{lccc}
\toprule
Prompt $\setminus$ in \% & Sensitivity & MAP & F$_1$ \\
\midrule
A & $\bm{36.92}$ & $\bm{26.38}$ & $\bm{23.75}$ \\
B & 35.54 & 26.00 & 23.75 \\
C \& D & 23.57 & 22.62 & 17.05 \\
\bottomrule
\end{tabular}
\label{tab:prompt_comparison}
\begin{tcolorbox}[notitle,boxrule=0pt,
boxsep=0pt,left=0.6em,right=0.6em,top=0.5em,bottom=0.5em,
colback=gray!10,
colframe=gray!10]
\scriptsize
\textbf{A} \& {\color{NavyBlue}{\textbf{(B)}}}: ``{\color{NavyBlue}{(System: You are an expert auditor with perfect knowledge of the IFRS accounting standard.)}} Out of all document segments provided below which ones are the 5 most relevant for fulfilling the IFRS requirement? {\color{NavyBlue}{(Think step by step. )}} IFRS requirement: \{requirement\} document segments: \{document\} Your answer should only contain the ids of the relevant document segments. Example: ['1129', '1139','1159', '1161', '829']. Your answer needs to be machine readable. Do not add any additional text.''\vspace{0.5em}
% \textbf{B}: ``System: You are an expert auditor with perfect knowledge of the IFRS accounting standard. Out of all document segments provided below which ones are the 5 most relevant for fulfilling the IFRS requirement? Think step by step: IFRS requirement: {requirement} document segments: {document} Your answer should only contain the ids of the relevant document segments. Example: ['1129', '1139','1159', '1161', '829']. Your answer needs to be machine readable. Do not add any additional text.''\vspace{0.5em}
% \textbf{C}: ``System: You are an expert auditor with perfect knowledge of the IFRS accounting standard. Out of all document segments provided below which ones are the 5 most relevant for fulfilling the IFRS requirement?Explain for each requirement why you selected the 5 most relevant requirements. Think step by step: IFRS requirement: {requirement} document segments: {document} Format your output complying to the following json schema: ['explanation': 'The most relevant document segments ...','answer': ['1129', '1139','1159', '1161', '829']. Ensure that 'answer' is it's own key in the json schema. Your answer needs to be machine readable.''\vspace{0.5em}

\textbf{C} \& {\color{NavyBlue}{\textbf{(D)}}}: ``System: You are an expert auditor with perfect knowledge of the IFRS accounting standard. Out of all document segments provided below which ones are the 5 most relevant for fulfilling the IFRS requirement? Explain for each requirement why you selected the 5 most relevant requirements. {\color{NavyBlue}{(Each should only be a sentence long.)}} Think step by step: IFRS requirement: \{requirement\} document segments: \{document\} Format your output complying to the following json schema: \{'explanation': 'The most relevant document segments ...', 'answer': ['1129', '1139','1159', '1161', '829']\}. Ensure that 'answer' is its own key in the json schema. Your answer needs to be machine readable.''
\end{tcolorbox}
\vspace{-0.5cm}
\end{table}

\subsection{Evaluation Metrics}
\label{sec:metrics}

To more accurately evaluate the performance of the four architectures, we employ the ``mean-average-precision'' (MAP) to comprehensively assess the performance of the recommender system across multiple queries. The average precision (AP) is calculated as
\begin{equation}
    \label{eq:mAP}
    \text{AP}=\frac{1}{\min(\{K, \text{A}(r)\})}\sum_{i=1}^{K}\left(\text{Precision}(i)\cdot \text{relevance}(i)\right)
\end{equation}
where $\text{relevance}(i)$ is $1$ if requirement $i$ is relevant and $0$ if it is not, $r$ is a given requirement, and $\text{A}(r)$ are the annotated text segments for requirement $r$. In addition, we deploy the custom performance metric sensitivity, which is a slightly modified version of recall and considers whether the relevant segments are contained in the set of recommendations (see \cite{Sifa19}).
% , designed to reflect user experience (see \cite{Sifa19}). 
Precision and sensitivity are defined as
\begin{equation*}
    \label{eq:precision_and_sensitivity}
    \text{Precision}(r) = \frac{|\text{P}(r,K) \cap \text{A}(r)|}{K},\;\text{Sensitivity}(r) = \frac{|\text{P}(r,K) \cap \text{A}(r)|}{\min(\{K,\text{A}(r)\})}
\end{equation*}
with $\text{P}(r,K)$ being the top $K$ recommended text segments for $r$.
For our experiments, we choose $K=5$, as this proves to deliver optimal results with respect to user experience. Finally, the conventional F$_1$-score (with respect to the top-$K$ predictions) is defined as the harmonic mean over precision and recall for a given requirement. %:
% \begin{equation}
%     \text{F}_{1}(K, r) = 2 \cdot \frac{\text{Precision}(K,r) \cdot \text{Recall}(K,r)}{\text{Precision}(K,r)+\text{Recall}(K,r)}.
% \end{equation}
All scores are averaged across the dataset to determine the overall performance.

\subsection{Evaluation and Results}

% \begin{table}
%  \caption{Test set results for the recommendation of relevant financial report segments for legal requirements of the IFRS accounting standard. ZeroShotALI outperforms all competing methods in one-shot-recall and F$_1$ score.}
%   \centering
%   \begin{tabular}{lcccc}
%     \toprule
%     Model & One Shot Recall & F$_1$  & Recall & Precision \\
%     \midrule
%     SentenceBERT & 0.81 & 0.28 & 0.41 & 0.52 \\
%     Chroma & 0.47 & 0.13 & 0.19 & 0.26 \\
%     Chroma + GPT-4 & 0.58 & 0.19 & 0.27 & 0.35 \\
%     ZeroShotALI & $\mat{0.83}$ & $\mat{0.31}$ & $\mat{0.45}$ & $\mat{0.58}$ \\
%     \bottomrule
%   \end{tabular}
%   \label{tab:my_label}
% \end{table}

\begin{table}
 \caption{Test set results for the top 5 recommendation of relevant financial report segments for legal requirements of the IFRS accounting standard. ZeroShotALI outperforms all competing methods in mean sensitivity, mean average precision (MAP) and F$_1$ score.}
  \centering
  \adjustbox{max width=\linewidth}{
    \begin{tabular}{lccc}
    \toprule
    Model $\setminus$ in \% & Sensitivity &  MAP & F$_1$  \\
    \midrule
    Chroma (Ada V1) & 14.00 & 7.12 & 9.12 \\
    Chroma (Ada V2) & 25.73 & 17.33 & 13.15 \\
    % Chroma (ADA V1) + GPT-3.5 Turbo & 14.95 & 11.67 & 7.54 \\
    Chroma (Ada V2) + GPT-3.5 Turbo & 29.95 & 21.32 & 15.74 \\
    Chroma (Ada V2) + GPT-4 & 35.30 & 24.72 & 18.53 \\
    SentenceBERT (from \cite{biesner2022zeroshot}) & 52.12 & 39.00 & 27.69 \\
    ZeroShotALI (this work) & $\mat{57.62}$ & $\mat{44.65}$ & $\mat{30.57}$ \\
    \bottomrule
    \end{tabular}}
  \label{tab:baseline_comparison}
\end{table}

% VERSION June 14th 12 pm
% \begin{table}
%  \caption{Test set results for the top 5 recommendation of relevant financial report segments for legal requirements of the IFRS accounting standard. ZeroShotALI outperforms all competing methods in mean sensitivity, mean average precision (MAP) and F$_1$ score.}
%   \centering
%   \begin{tabular}{lccc}
%     \toprule
%     Model $\setminus$ in \% & Sensitivity &  MAP & F$_1$  \\
%     \midrule
%     SentenceBERT & 52.12 & 39.00 & 27.69 \\
%     Chroma & 25.73 & 17.33 & 13.15 \\
%     Chroma + GPT-4 & 35.30 & 24.72 & 18.53 \\
%     ZeroShotALI & $\mat{57.62}$ & $\mat{44.65}$ & $\mat{30.57}$ \\
%     \bottomrule
%   \end{tabular}
%   \label{tab:my_label}
% \end{table}

% \begin{table}
%  \caption{Test set results for the recommendation of relevant financial report segments for legal requirements of the IFRS accounting standard. ZeroShotALI outperforms all competing methods in one-shot-recall and F$_1$ score.}
%   \centering
%   \begin{tabular}{lcccc}
%     \toprule
%     Model $\setminus$ in \% & One Shot Recall & F$_1$  & Recall & Precision \\
%     \midrule
%     SentenceBERT & 80.82 & 27.69 & 41.05 & 31.48 \\
%     Chroma & 47.34 & 13.15 & 19.86 & 16.42 \\
%     Chroma + GPT-4 & 58.03 & 18.53 & 26.82 & 22.13 \\
%     ZeroShotALI & $\mat{83.43}$ & $\mat{30.57}$ & $\mat{45.47}$ & $\mat{34.67}$ \\
%     \bottomrule
%   \end{tabular}
%   \label{tab:my_label}
% \end{table}
As depicted in Table \ref{tab:baseline_comparison}, the performance analysis reveals that the two architectures based on SentenceBERT surpass the vector store-based architectures. Notably, ZeroSHotALI, combining SentenceBERT with GPT-4, demonstrates the highest performance which can be attributed to several factors. %achieving a remarkable one-shot-recall score of 83\%.
% The superior performance of this combined architecture can be attributed to several factors.

Firstly, the vector store-based architectures rely on embeddings from generically pre-trained language models that exhibit no domain specific fine-tuning and leverage approximate nearest-neighbor calculations to return text matches. In contrast, the SentenceBERT model was fine-tuned specifically for the task of retrieving semantically similar text passages within an auditing context. This custom training process enables SentenceBERT to capture the intricacies and nuances of the auditing domain, leading to more accurate and contextually relevant retrievals. The observed subpar performance of vector store-based architectures in our use cases raises an important question regarding the suitability of such systems in various applications. Currently, many applications leverage retrieval-augmented generation using vector databases, assuming that these systems can provide reliable results. However, our findings suggest that incorporating domain-specific, fine-tuned retrieval systems like SentenceBERT could significantly enhance the performance of such applications.

%\section{Methodology}
%Introduction paragraph.
%\subsection{Generic Text Matching}
%\subsection{Contradiction Detection}
%\subsection{Completeness Check}

%\section{Experiments}
%Introduction paragraph.
%\subsection{Generic Text Matching}
%\subsubsection{Data}
%\subsubsection{Results}
%\subsection{Contradiction Detection}
%\subsubsection{Data}
%\subsubsection{Results}
%\subsection{Completeness Check}
%\subsubsection{Data}
%\subsubsection{Results}

% \begin{table}[t]
% \small
% \centering
% \caption{Caption}
% \label{tab:model_results}
% \input{tables/model_results.tex}
% \end{table}

\section{Conclusion and Future Work}
\label{section:conclusion}

In this study, we conducted an extensive evaluation to explore the viability of employing a "frozen" LLM for auditing financial documents. Our assessment focused on implementing various architectures tailored to match the text of IFRS-compliant annual reports with legal requirements. The results of our analysis revealed that our proposed system, ZeroShotAli, which combines a domain-tuned SentenceBERT model with OpenAI’s GPT4, outperforms other systems significantly. This finding underscores the advantage of deploying domain-specific solutions when in-domain data is available, surpassing generic systems such as vector databases with general purpose embeddings. Our study accentuates the potential for substantial performance gains through the utilization of domain-specific approaches in the context of retrieval-augmented generation. This study identifies three key areas for future research: (1) deploying domain-specific fine-tuned LLMs for auditing purposes using open-source models like LLaMA, (2) exploring advanced prompt tuning methodologies such as Chain-of-Thought \cite{wei2022chain} or Tree-of-Thoughts \cite{yao2023tree} approaches, and (3) expanding the ZeroShotAli system to assess requirement completeness based on relevant text passages.
% (1) deploying domain-specific fine-tuned LLMs for auditing purposes using open-source models like LLaMA, (2) exploring advanced prompt tuning methodologies such as Chain-of-Thought \cite{wei2022chain} or Tree-of-Thoughts \cite{yao2023tree} approaches, and (3) expanding the ZeroShotAli system to assess requirement completeness based on relevant text passages.

\begin{acks}
This research has been partially funded by the Federal Ministry of Education and Research of Germany and the state of North-Rhine Westphalia as part of the Lamarr-Institute for Machine Learning and Artificial Intelligence.
% This research has been funded by the BMBF and state of NRW as part of the Lamarr-Institute.
\end{acks}

\bibliographystyle{ACM-Reference-Format}
\bibliography{bibliography}  % sample-base

% \appendix



\end{document}